# FULLY AUTOMATIC COMPUTER-AIDED MASS DETECTION AND SEGMENTATION VIA PSEUDO-COLOR MAMMOGRAMS AND MASK R-CNN

*Hang Min, Devin Wilson, Yinhuang Huang, Siyu Liu, Stuart Crozier, Andrew P Bradley, Shekhar S. Chandra*

School of Information Technology and Electrical Engineering, University of Queensland, Australia.

## ABSTRACT

Mammographic mass detection and segmentation are usually performed as serial and separate tasks, with segmentation often only performed on manually confirmed true positive detections in previous studies. We propose a fully-integrated computer-aided detection (CAD) system for simultaneous mammographic mass detection and segmentation without user intervention. The proposed CAD only consists of a pseudo-color image generation and a mass detection-segmentation stage based on Mask R-CNN. Grayscale mammograms are transformed into pseudo-color images based on multi-scale morphological sifting where mass-like patterns are enhanced to improve the performance of Mask R-CNN. Transfer learning with the Mask R-CNN is then adopted to simultaneously detect and segment masses on the pseudo-color images. Evaluated on the public dataset INbreast, the method outperforms the state-of-the-art methods by achieving an average true positive rate of 0.90 at 0.9 false positive per image and an average Dice similarity index of 0.88 for mass segmentation.

## 1. INTRODUCTION

Breast cancer is one of the most frequently diagnosed cancers among women worldwide [1]. Mammography is the primary tool for breast screening, where the large volume of data generated can lead to fatigue and missed detections in human analysis. To assist radiologists and increase confidence of detection, computer-aided detection (CAD) systems have been developed as a 'second pair of eyes' in mammography interpretation [2].

A breast mass CAD system usually extracts a number of region candidates that may contain masses firstly, and then classifies the region candidates as abnormal or normal based on the features extracted from these regions. Traditional CADs mainly rely on unsupervised region candidate generation and hand-crafted features to detect breast masses [2]. However, it can be difficult to achieve a balance between discriminative power and robustness when using hand-crafted features. Recent developments in deep learning (DL) based methods can provide more robust solutions to this problem. These methods utilize convolutional neural networks (CNNs) to learn meaningful features directly from the training data and have achieved promising results. Dhungel et al. [3] adopted a multi-scale deep belief network and Gaussian mixture model for region candidate proposal, and then classified the region candidates with combinations of R-CNN and random forests. Li et al. [4] adopted an unsupervised region proposal method based on morphological analysis and a CNN to classify the region candidates. However, these CADs only localize masses without segmenting them. To segment masses, a separate segmentation stage needs to be added to the system. Dhungel et al. [5] extended their previous work [3] by adding a segmentation scheme based on conditional random field and a level set method. However, this work only performs segmentation on true positive (TP) detections and the false positive (FP) detections need to be manually rejected. Oliveira et al. [6] used the residual neural network (ResNet) to generate region candidates, a CNN to classify the candidates and a segmentation refinement method to generate the contour of masses. Both Oliveira et al. [6] and Dhungel et al. [5] treat mass detection and segmentation as separate tasks and contain multiple DL networks which need to be tuned sequentially. Currently, there is still a need for a more integrated framework that can perform mass detection and segmentation at the same time.

In this work, we propose a fully-integrated mammographic CAD system that can detect and segment masses simultaneously without user intervention in a simple framework. The system only consists of two major stages which are pseudo-color image generation and detection-segmentation based on the Mask R-CNN [7] as shown in Figure 1. A key contribution of this work is that we introduce the concept of pseudo-color mammogram (PCM) which renders mass-like patterns with color contrast with respect to the background. The PCM is generated by appending two morphologically filtered mammograms to the grayscale mammogram in two adjacent image channels as shown in Figure 1. To identify masses on PCMs, we adopt transfer learning with Mask R-CNN due to the limited size of publicly available mammographic datasets. The Mask R-CNN is a recently proposed general framework for object detection and segmentation [7]. The proposed pseudo-color scheme paired with the Mask R-CNN deep learning framework provides an integrated solution to mammographic mass detection and segmentation that does not require any manual intervention or hand-crafted features. This work is evaluated on the publicly available INbreast dataset [8] and outperforms the state-of-the-art methods on identical evaluation sets. The source code for this work has been made available online [9].

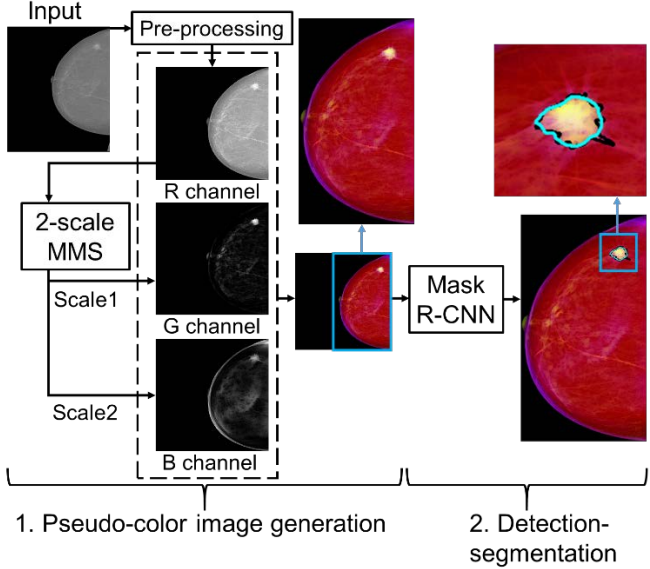

Figure 1. The diagram of the proposed CAD. MMS stands for multi-scale morphological sifting. The black outlines represent the ground truth of the mass and the cyan outlines represent the segmentation generated by the proposed method.

## 2. METHODOLOGY

### 2.1. Data

The INbreast dataset [8] is used for evaluating the proposed method. This dataset is currently the largest publicly available full-digital mammographic dataset with mammograms precisely annotated [5]. It contains 115 cases with 410 mammograms. There are 116 breast masses in total, within the size range of $[15mm^2, 3689mm^2]$. The pixel size of the mammograms is $70\mu m$, and the bit depth is 14-bit. For the evaluation on INbreast, we use the same validation data partition as [5, 10] to enable a direct comparison, where the dataset is randomly split into 60% for training, 20% for validation and 20% for testing five times.

### 2.2. Pre-processing

The breast region is extracted by thresholding and the redundant background is cropped away [10]. The mammogram is then normalized to 16-bit and padded into a square shape. To speed up the process, the mammogram is sub-sampled to 1/4 of its original size using the low-pass component of a two-level Daubechies 2 wavelet transform.

### 2.3. Pseudo-color mammogram generation

In this stage, the mammogram is transformed into a pseudo-color mammogram (PCM) to enhance the mass-like structures. The grayscale mammogram (GM) is put in the first channel, and the other two channels are filled with two images generated by the multi-scale morphological sifter (MMS) [10] to provide complementary information as shown in Figure 1. The MMS uses morphological filters with oriented linear structuring elements (LSEs) to extract mass-like patterns including linear spicules that are normally present in breast masses [10].

The MMS is described in Eq. (1), where an input image $F$ is processed by two sets of morphological filters, $L(M_1(i), \theta(n))$ and $L(M_2(i), \theta(n))$ $(n = 0,1, \ldots N-1)$ on scale $i$ $(i = 1, \ldots, I)$. Here, '∘' stands for morphological opening. Each set of filters contains $N$ LSEs. $M_1$ and $M_2$ stand for the magnitude, and $\theta(n) = n \times (180°/N)$ stands for the orientation of the $n^{th}$ LSE in each set. On scale $i$, the MMS is able to extract patterns whose diameter is within the range of $[M_1(i), M_2(i)]$. Given the area range of the target for detection $[A_{min}, A_{max}]$, the magnitudes $M_1$ and $M_2$ are defined in Eq. (2, 3), where $P$ is the pixel size of the original image and $S$ is the resizing factor in the pre-processing stage.

$$MMS = \sum_{n=0}^{N-1}\{F - [F \circ L(M_2(i), \theta(n))]\} \circ L(M_1(i), \theta(n)) \quad (1)$$

$$M_1(i) = \frac{2}{P \cdot S}\left(\frac{A_{min}}{\pi}\right)^{0.5} \cdot \left(\frac{A_{max}}{A_{min}}\right)^{\frac{0.5(i-1)}{I}} \quad (2)$$

$$M_2(i) = \frac{2}{P \cdot S}\left(\frac{A_{min}}{\pi}\right)^{0.5} \cdot \left(\frac{A_{max}}{A_{min}}\right)^{\frac{0.5i}{I}} \quad (3)$$

In this work, two scales are used for the MMS ($I = 2$), the size range of the masses $[A_{min}, A_{max}]$ has been suggested in [8] as $[15mm^2, 3689mm^2]$, and the number of LSEs ($N$) in each scale is set as the default value 18 as in [10]. The resizing factor $S = 4$ as mentioned in the pre-processing stage. The two output images and the GM are linearly scaled to 8-bit. A PCM consists of the GM in the R (red) channel, the output image of MMS from scale 1 in the G (green) channel and the output image from scale 2 in the B (blue) channel as shown in Figure 2. If a mass is relatively small and lands in the size range of scale 1, it will have a higher intensity in the G channel and appears to be more yellow on the PCM as shown in Figure 2 (a). If a mass is relatively large and lands in the size range of scale 2, it will have a higher intensity in the B channel and appears to be purple on the PCM as shown in Figure 2 (b). This pseudo-color rendering scheme can enhance mass-like patterns by adding color contrast between the mass and the background when combining three channels.

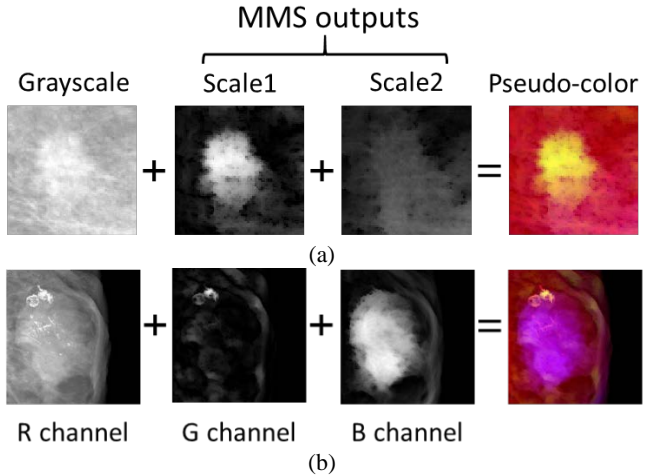

Figure 2. Pseudo-color enhancement for masses in different sizes. (a) displays a small mass with an equivalent diameter of 6 mm and (b) displays a large mass with a diameter of 27 mm.

## 2.4. Applying Mask R-CNN

In this work, we adopt transfer learning with a pre-trained Mask R-CNN model, since the mammographic dataset is limited in size. The Mask R-CNN provides a general framework for simultaneous mass detection and segmentation. It consists of the Faster R-CNN for object detection and a fully convolutional network (FCN) for a pixel-to-pixel segmentation. The Faster R-CNN uses a region proposal network to propose bounding box region candidates and then classifies these candidates into different categories. The FCN runs in parallel to perform segmentation on the region candidates. The multi-task loss function of Mask R-CNN is $L = L_{cls} + L_{bbox} + L_{msk}$, where $L_{cls}$ is the classification loss and $L_{bbox}$ is the bounding box regression loss [7]. The mask loss $L_{msk}$ is defined as the binary cross-entropy loss with a per-pixel sigmoid activation [7].

The Mask R-CNN training is initialized using the 'mask_rcnn_balloon' pre-trained model which was trained for a binary classification problem (separating balloons from the background) [11]. The ResNet101 is used as the Mask R-CNN backbone. The image resizing mode in Mask R-CNN is set to 'square' and images are resized into $1024 \times 1024$. Images in training and validation sets are augmented in one way randomly chosen from flipped up/down, left/right and rotated in $90°, 180°, 270°$. All layers in the network are then trained through 10 epochs, with 100 training and 10 validation steps in each epoch, which takes around 10 minutes. The training is repeated 5 times and the model with the lowest sum of training and validation loss among all epochs is selected for testing. Except for the parameter settings mentioned above, the rest of the parameters all remain the same as the default values in [11]. A comparison experiment using GMs as the input of Mask R-CNN is also carried out. The pseudo-color image generation is implemented in MATLAB 2018b and computed on a Dell desktop with Intel Core i7-4790CPU@3.60 GHz, 16 GB RAM. The Mask R-CNN is implemented in Python 3 and ran on a Dell EMC PowerEdge R740 server with 384GB DDR4 RAM and two NVIDIA Tesla V100 16GB.

## 3. RESULTS

The free response operating characteristic (FROC) curves of the testing performance using PCMs and GMs with Mask R-CNN are shown in Figure 3. A detection is regarded as a true positive (TP) if it has a Dice similarity index (DSI) no less than 0.2 to the ground truth [12]. Using PCMs with Mask R-CNN achieves an average true positive rate (TPR) of $0.90 \pm 0.05$ at 0.9 false positive per image (FPI) as marked with a red dot on Figure 3. The partial area under the FROC curve (AUFC) [13] is 0.90 in the FPI range of $[0, 5]$. The average DSI for mass segmentation is $0.88 \pm 0.10$. Using GMs and Mask R-CNN yields an average TPR of $0.90 \pm 0.05$ with a FPI of 1.9 as marked with a blue dot on Figure 3. The AUFC is 0.85 and the average DSI is $0.87 \pm 0.09$. Several detection and segmentation examples on PCMs are shown in Figure 4. Table 1 shows the overall performance comparison between the proposed methods and several previous methods.

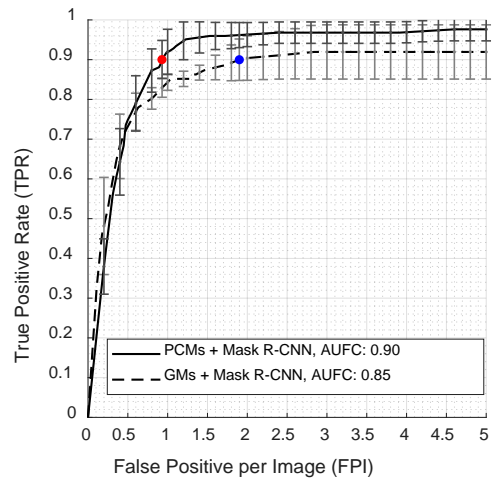

Figure 3. FROC curves for the proposed method using PCMs and GMs with Mask R-CNN. The red dot marks an average TPR of 0.9 at 0.9 FPI using PCMs and the blue dot marks an average TPR of 0.9 at 1.9 FPI using GMs.

## 4. DISCUSSION

The PCM generation provides a new way of generating multi-channel inputs for DL networks such as the Mask R-CNN. It selectively enhances mass-like patterns with color contrast as shown in Figure 2. Li et al. [4] also attempted to use a multi-channel input containing the original, a gradient and a texture image, for CNN networks in mass detection. However, this study stated that the multi-channel input did not improve the performance of CNN since the gradient and texture filters could be learned by CNN directly through convolutional feature maps. The MMS uses morphological filters which cannot be directly learned by CNN without adaptations to the convolution layer [14]. Therefore, the usage of the PCMs generated by MMS demonstrates the ability to improve the detection performance of Mask R-CNN compared with only using GMs as shown in Figure 3 and Table 1.

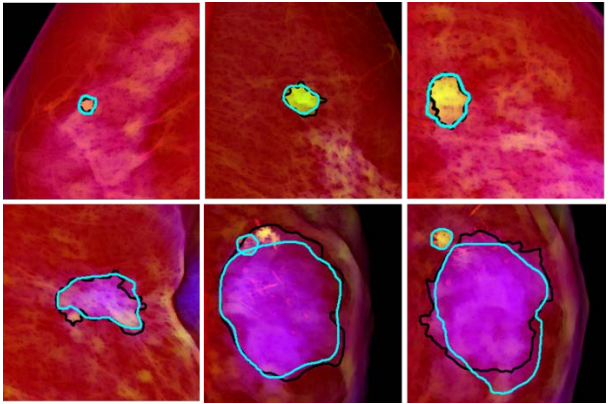

Figure 4. Detection and segmentation examples of masses in different shapes and sizes using PCMs and Mask R-CNN. The black lines represent the ground truth of the masses. The cyan lines represent the segmentation of the detected regions.

Table 1. Mass detection and segmentation performance comparison between PCMs and GMs + Mask R-CNN and several previous methods.

| Method | Avg. TPR @ FPI | DSI |
|---|---|---|
| PCMs+Mask R-CNN | 0.90 ± 0.05 @ 0.9 | 0.88 ± 0.10 |
| GMs+Mask R-CNN | 0.90 ± 0.05 @ 1.9 | 0.87 ± 0.09 |
| Min et al. [10] | 0.90 ± 0.06 @ 0.9 | 0.86 ± 0.08 |
| Dhungel et al. [5] | 0.90 ± 0.02 @ 1.3 | 0.85 ± 0.02 |
| Oliveira et al. [6] | 0.85 ± 0.07 @ 3±0.2 | 0.83 ± 0.10 |
| Li et al. [4] | 0.90 @ 1.88 | - |

The Mask R-CNN integrates the region candidate generation, feature extraction and region classification stages in one pipeline. It performs mass detection and segmentation tasks altogether and the network can be optimized as a whole. The proposed CAD does not require massive training of the Mask R-CNN. Each round of training only takes approximately 10 minutes with 10 epochs. The testing of the Mask R-CNN model takes less than 1s per image.

Table 1 shows that the proposed CAD using PCMs and Mask R-CNN achieves a similar detection performance and a higher segmentation performance compared to [10], and outperforms [5, 6] in both detection and segmentation on INbreast data. Dhungel et al. [5] and Oliveira et al. [6] both have separate networks for region candidate proposal, classification and mass segmentation, which need to be tuned sequentially, while the proposed CAD integrates all these stages. This work also does not require users to reject the FPs before performing segmentation as in Dhungel et al. [5]. It is uncertain if our performance is more favorable to Li et al. [4], since this study was evaluated on a private dataset. However, Li et al. [4] is a detection system that does not segment masses.

There are, however, limitations in this work. The variation in DSI is relatively high as shown in Table 1. With a larger dataset, the performance stability can potentially be improved. This CAD also does not perform mass characterization. In future work, we would like to evaluate our method on larger mammographic datasets, and investigate the method's capability of identifying different types of lesions.

## 5. CONCLUSION

In this work, we proposed an integrated mammographic CAD for simultaneous mass detection and segmentation based on pseudo-color mammograms and Mask R-CNN. The novel pseudo-color image generation stage based on MMS can provide color contrast between the masses and the background tissue, which significantly improves the detection performance of Mask R-CNN compared to using grayscale mammograms. The proposed CAD system does not require hand-crafted features or user intervention. Compared with the state-of-the-art methods, the system achieves favorable performance in both mass detection and segmentation in a simple framework.